\documentclass[10pt,twocolumn,letterpaper]{article}

\usepackage[pagenumbers]{cvpr}
\usepackage{multirow}
\usepackage{svg}
\usepackage{booktabs}
\usepackage{makecell}
\usepackage{placeins}
\usepackage{xcolor} % 导入颜色包

\definecolor{mygreen}{rgb}{0.0, 1.0, 0.5}
\definecolor{cvprblue}{rgb}{0.21,0.49,0.74}
\usepackage[pagebackref,breaklinks,colorlinks,allcolors=cvprblue]{hyperref}

\def\paperID{6670} % *** Enter the Paper ID here  6670
\def\confName{CVPR}
\def\confYear{2026}

\title{S$^2$-MLLM: Boosting Spatial Reasoning Capability of MLLMs for 3D Visual Grounding with Structural Guidance}

\author{
Beining Xu$^{1}$\thanks{These authors contributed equally.} ,
Siting Zhu$^{1}$\footnotemark[1],
Zhao Jin$^{2}$,
Junxian Li$^{1}$,
Hesheng Wang$^{1}$\thanks{Corresponding Author} \\
\small{$^1$Shanghai Jiao Tong University, $^2$Nanyang Technological University}
}

\begin{document}
%\twocolumn[{
%\renewcommand\twocolumn[1][]{#1}

\maketitle

\begin{abstract}
3D Visual Grounding (3DVG) focuses on locating objects in 3D scenes based on natural language descriptions, serving as a fundamental task for embodied AI and robotics.
Recent advances in Multi-modal Large Language Models (MLLMs) have motivated research into extending them to 3DVG.
However, MLLMs primarily process 2D visual inputs and struggle with understanding 3D spatial structure of scenes solely from these limited perspectives. Existing methods mainly utilize viewpoint-dependent rendering of reconstructed point clouds to provide explicit structural guidance for MLLMs in 3DVG tasks, leading to inefficiency and limited spatial reasoning. 
To address this issue, we propose S$^2$-MLLM, an efficient framework that enhances spatial reasoning in MLLMs through implicit spatial reasoning. 
We introduce a spatial guidance strategy that leverages the structure awareness of feed-forward 3D reconstruction. 
By acquiring 3D structural understanding during training, our model can implicitly reason about 3D scenes without relying on inefficient point cloud reconstruction.
Moreover, we propose a structure-enhanced module (SE), which first employs intra-view and inter-view attention mechanisms to capture dependencies within views and correspondences across views.
The module further integrates multi-level position encoding to associate visual representations with spatial positions and viewpoint information, enabling more accurate structural understanding. 
Extensive experiments demonstrate that S$^2$-MLLM unifies superior performance, generalization, and efficiency, achieving significant performance over existing methods across the ScanRefer, Nr3D, and Sr3D datasets.
Code will be available upon acceptance.
%Code will be released at \url{http://www.pamitc.org/documents/mermin.pdf}.
\end{abstract} 

\begin{figure}[t]
    \centering
    \vspace{-0.5cm}
    \includegraphics[width=1.0\linewidth]{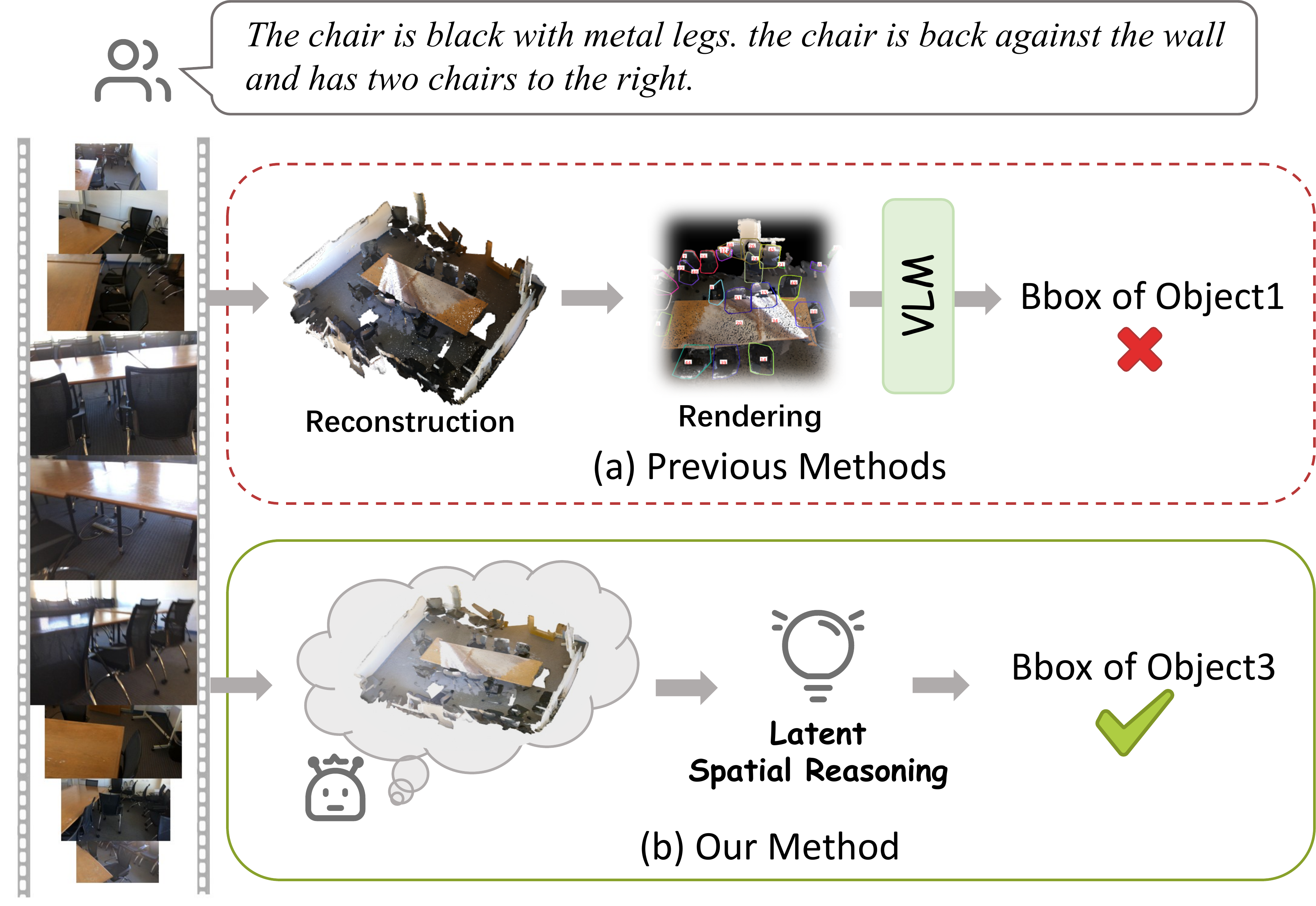}
    \vspace{-0.5cm}
    \caption{Comparison of previous methods and our method.
(a) Previous methods typically reconstruct point clouds of 3D scenes explicitly and then render 2D images to obtain structure guidance. 
(b) Our method leverages spatial guidance to understand the 3D structure during training, allowing the model to perform implicit spatial reasoning in the latent space without requiring point-cloud reconstruction at inference.
    }
    \label{fig:comparison}
\end{figure}

\section{Introduction}
\label{sec:intro}
3DVG aims to locate referred objects in 3D scenes based on textual descriptions~\cite{chen2020scanrefer,achlioptas2020referit3d,jin2023context}, serving as a key capability for embodied AI~\cite{lu2024scaneru, zhu2024sni, zhu2025sni, zhu2024semgauss} and augmented reality~\cite{liu2024t3dvgSurvey, xu2025sgloc, sha2024towards, zhu20243d}.
Compared to 2D visual grounding, 3DVG demands a thorough understanding of spatial relationships and 3D scene structures~\cite{yang2023exploiting,liu2025survey,hong20233d,jia2024sceneverse,wang20233drp}, thereby posing greater challenges.

Traditional approaches~\cite{achlioptas2020referit3d,cai20223djcg,chen2022language,huang2022multi,guo2025tsp3d,jain2022bottom,peng2025proxytransformation,wang2025liba,wu2023eda,yuan2021instancerefer,zhao20213dvg,guo2023viewrefer} primarily rely on training within a limited scope of datasets, resulting in limited generalization and scalability in real-world scenes~\cite{li2025seeground,jin2025spazer}.
To overcome this limitation, leveraging MLLMs~\cite{wang2024qwen2,achiam2023gpt,li2024llava, xu2024vlm,li2025chemvlm, li2025faithact} for 3DVG through their reasoning and generalization abilities has emerged as a promising direction.
However, a significant gap remains between the 2D-centric training of MLLMs and spatial understanding capabilities demanded by 3DVG~\cite{li2025seeground,jain2025unifying,zhan2025actial}. 
MLLMs are primarily trained on extensive image-text data without exposure to the 3D physical world~\cite{qi2025gpt4scene,ma2024llms}, which involves additional factors such as depth, viewpoint, layout, and structure~\cite{liu2025survey,liu2024t3dvgSurvey}.
Therefore, MLLMs are incapable of understanding 3D scenes solely from 2D images~\cite{zheng2025video,hong20233d}.

Existing approaches~\cite{li2025seeground,jin2025spazer,qi2025gpt4scene} attempt to compensate for this limitation by explicitly reconstructing point clouds of 3D scenes. 
These reconstructed point clouds are then rendered into 2D representations that preserve the layout and spatial relations, such as multi-view images~\cite{li2025seeground,jin2025spazer} or Bird's Eye View (BEV) images~\cite{qi2025gpt4scene}.
MLLMs can directly process these rendered images to select target objects.
However, rendered images from specific viewpoints are unable to reflect the comprehensive structure of 3D scenes~\cite{Tulsiani_2018_ECCV, Liu_2020_NeuralSparseVoxelFields} and are inevitably affected by viewpoint selection as well as occlusion.
Moreover, these methods need to reconstruct point clouds explicitly during inference, leading to low efficiency.

To address these challenges, we propose a novel framework S$^2$-MLLM, which enhances the spatial reasoning of MLLMs for 3DVG.
As shown in Fig.~\ref{fig:comparison}, our key insight is to encourage the model to implicitly internalize 3D structure awareness during training.
This design enables our S$^2$-MLLM to reason implicitly about 3D scenes within the latent feature space, without requiring extra reconstruction or rendering at inference.
Specifically, we propose to introduce spatial guidance by leveraging the capability of feed-forward 3D 
reconstruction~\cite{yang2025fast3r,wang2024dust3r,wang2025vggt}.
These reconstruction techniques can directly derive 3D structures from multi-view inputs, demonstrating an inherent capability for spatial understanding~\cite{zhang2025advances,wu2025spatial}.
Building upon this property, we integrate the reconstruction objective into the training pipeline through end-to-end joint optimization, enabling the model to learn structure-aware visual representations and spatial reasoning capability.

Moreover, we propose a structure-enhanced (SE) module to further enhance the spatial understanding of MLLMs from the perspectives of position and viewpoint.
We observe that (i) without explicit association between position cues (\eg, 3D coordinates, camera rays) and visual appearance, MLLMs struggle to understand fine-grained spatial relations such as distance, direction, and relative relations; (ii) MLLMs are pretrained on independent image–text pairs, making it difficult to maintain semantic consistency across views.
Therefore, we first integrate a multi-level position encoding to enhance the modeling of position and viewpoint information.
Secondly, we employ inter-view attention mechanism to enforce semantic alignment during viewpoint transitions across views. 
Within each view, we conduct intra-view attention to capture dependencies between patches, which improves local and global context understanding. %improving local and 
We conduct extensive experiments on both in-domain~\cite{chen2020scanrefer,achlioptas2020referit3d} and out-of-domain benchmarks~\cite{baruch2021arkitscenes,mao2022multiscan}.
S$^2$-MLLM achieves outstanding performance in terms of accuracy, efficiency, and generalization.
This trade-off enables practical deployment in real-world applications and embodied robotics.

Overall, we provide the following contributions:
\begin{itemize}
    \item We propose S$^2$-MLLM, an effective framework that enhances the spatial reasoning of MLLMs, thereby improving performance for 3DVG.
    \item  We introduce a \textbf{spatial guidance} strategy that utilizes the capability of feed-forward 3D reconstruction to encourage our model to perform latent spatial reasoning. 
    \item We design a \textbf{structure-enhanced module (SE)} module that enhances spatial understanding through modeling position and viewpoint, as well as employs intra-view and inter-view attention to strengthen contextual alignment and cross-view consistency.
    \item Extensive experiments on \textit{ScanRefer}, \textit{Nr3D}, and \textit{Sr3D} datasets demonstrate that S$^2$-MLLM significantly outperforms baselines.
    Multiple out-of-domain evaluations indicate the generalization ability of S$^2$-MLLM. 
\end{itemize}

\section{Related Work}
\label{sec:related_work}
\subsection{Supervised 3D Visual Grounding}
3DVG is the task of 3D object localization in points or RGB-D scans via natural language descriptions~\cite{achlioptas2020referit3d,cai20223djcg,chen2022language,huang2022multi,guo2025tsp3d,jain2022bottom,peng2025proxytransformation,wang2025liba,wu2023eda,yuan2021instancerefer,zhao20213dvg}.
Existing traditional methods typically adopt a fully supervised paradigm, which can be categorized into two-stage methods and one-stage methods depending on their network architectural designs.
Two-stage methods~\cite{achlioptas2020referit3d,chen2020scanrefer,chen2022d,chen2022language,yuan2021instancerefer,zhao20213dvg,zhang2024multi,chang2024mikasa} follow a proposal–matching pipeline: 3D object proposals are obtained from pretrained detector or segmentor~\cite{qi2019deep,jiang2020pointgroup,schult2022mask3d} and then matched with the query.  
In recent years, more methods~\cite{yang2021sat,huang2022multi,guo2023viewrefer,miyanishi2024cross3dvg,zhu2024unifying,bakr2022look} integrates multi-modal information, such as 2D images and multi-view contexts.
To avoid relying on pre-trained proposal generators, one-stage methods~\cite{jain2022bottom,luo20223d,guo2025tsp3d,wu2023eda} directly regress the 3D box by densely aligning language with point-level features. 
Although these methods have achieved impressive accuracy on public benchmarks, they often suffer from limited generalization when applied to real-world 3D scenes.

\subsection{Zero-shot 3D Visual Grounding}
The reasoning and generalization abilities of LLMs have motivated several studies to explore zero-shot 3DVG.
These approaches typically decompose the 3DVG task into a sequence of sub-tasks that can be processed by LLMs, reducing the reliance on large-scale and high-quality 3D annotations.
However, as LLMs cannot directly process 3D information, the modality gap remains a fundamental challenge.
Early approaches \cite{xu2024vlm,kerr2023lerf,yuan2024visual,yang2024llm} convert object attributes into textual descriptions, then use the reasoning ability of LLMs to choose the object that best matches the query.
This design completely ignores the importance of scene-level context, which is essential for 3DVG.
VLM-based (Vision Language Models) methods \cite{li2025seeground,jin2025spazer,zantout2025sort3d} mainly employ VLMs for query analysis, viewpoint selection, and target object selection.
Complex structures and spatial relations in 3D scenes cannot be captured solely from textual descriptions or images from specific viewpoints, which causes these methods to fail in situations that require spatial reasoning.

\subsection{MLLMs for 3D Scene Understanding}
MLLMs have also been recently extended to 3D scene understanding~\cite{chandhok2024scenegpt,zhi2025lscenellm}, enabling the construction of a generalist model that is capable of handling multiple 3D tasks, including 3DVG~\cite{hong20233d,yu2025inst3d,fu2024scene,zhu20233d}.
To enable LLMs to understand 3D scenes, early approaches~\cite{chen2024ll3da,guo2023point}  integrate the point cloud encoder with LLMs, relying on large-scale training to enhance LLMs' ability to process point clouds.
~\cite{chen2024grounded, huang2023embodied,wang2023chat,zhang2024chatscene} incorporate object-centric 3D representations that further improves performance.
However, point-cloud–based approaches suffer from a large modality gap and limited 3D annotations~\cite{qi2025gpt4scene,xu2024vlm}.
These limitations motivate a shift toward learning 3D scenes from 2D image sequences.
Llava-3d~\cite{zhu2024llava} projects multi-view CLIP features into 3D voxels and aggregates them to recover coarse 3D scene structure.
Recent methods~\cite{zheng2025video,wu2025spatial,zheng2025learning} instead treat 3D scenes as dynamic video sequences.
GPT4Scene~\cite{qi2025gpt4scene} relies on the BEV image rendered from reconstructed point clouds to obtain global information, while ROSS3D~\cite{wang2025ross3d} introduces cross-view and BEV generation tasks to encourage understanding of the layout.
However, the former is time-consuming during inference due to reconstruction, and the latter is easily influenced by obstructions and noise, making both unreliable for learning consistent 3D structures.

\section{Methods}
\label{sec:method}

\begin{figure*}
  \centering
  \vspace{-0.5cm}
  \includegraphics[width=\linewidth]{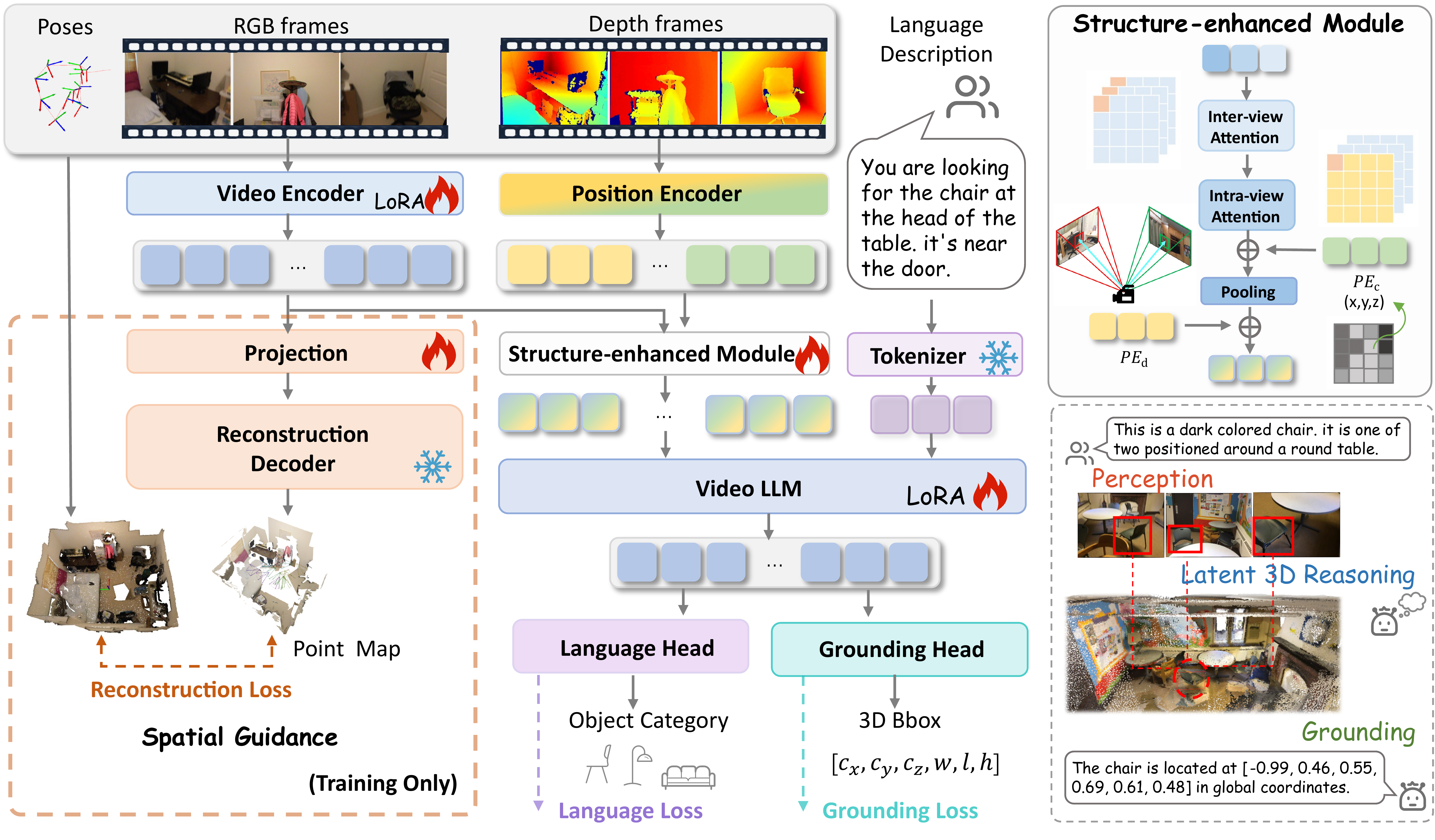}
    \vspace{-0.5cm}
  \caption{The Framework of S$^2$-MLLM. Our model takes sampled multi-view RGB-D frames, related camera parameters, language descriptions, and the bounding boxes of object proposals as inputs.
The shared video encoder and position encoder extract visual and geometric features from RGB-D frames.
The structure-enhanced module (SE) integrates visual features and position information to form the visual input for the LLM.
The Video LLM jointly processes the visual input and the tokenized query, enabling cross-modal understanding and reasoning.
The grounding head predicts the 3D bounding box (Bbox) of the target object. 
The language head generates the category of the target object.
while the reconstruction decoder predicts the point map to provide reconstruction supervision. 
Notice that the reconstruction is used only at training time.
    }
  \label{fig:framework}
\end{figure*}

We propose a task-specific model for 3D visual grounding based on a pre-trained MLLM~\cite{zhang2024video}, which effectively integrates information from texts and image sequences
Our method represents 3D scenes as video sequences, which preserves abundant texture and semantic cues of 2D images. 
Given multi-view RGB-D images $\{ (I_v, D_v) \}_{v=1}^{V}$, related camera parameters of 3D scene with candidate objects $\{o_i\}_{i=1}^N$, and a natural language description $Q$, our model predicts the 3D bounding box and category of target object $\hat{o}$ that best matches the description. 
The overview of our method is shown in Fig~\ref{fig:framework}.

In the following, we describe our framework in detail. 
Sec.~\ref{sec:Reconstruction} introduces the spatial guidance strategy.
Sec.~\ref{sec:Attention} details the intra-view and inter-view attention mechanisms.
Sec.~\ref{sec:position} presents the multi-level position encoding.
Finally, Sec.~\ref{sec:Loss} describes the overall training objectives.

\subsection{Spatial Guidance Strategy}
\label{sec:Reconstruction}
MLLMs are trained on massive collections of images, text, and videos, enabling them to integrate visual and textual information effectively~\cite{wu2023multimodal, yin2024survey}. 
However, the 2D visual priors encoded in MLLMs are insufficient for inferring 3D structure from RGB images~\cite{zheng2025video}, including global layout, geometric properties, cross-view correspondences, and fine-grained spatial relations.
The lack of 3D structure understanding restricts MLLMs from understanding the 3D physical world~\cite{ma2024llms,zha2025enable}.
Existing methods rely on reconstructing point clouds to produce BEV inputs~\cite{qi2025gpt4scene} or rendered images~\cite{li2025seeground,jin2025spazer} to provide structure guidance for MLLMs.
However, rendered images are easily influenced by the selection of viewpoints and occlusions~\cite{Tulsiani_2018_ECCV, Liu_2020_NeuralSparseVoxelFields}.
For example, relative spatial relations may change when observed from different viewpoints.
Consequently, relying solely on view-specific observations often leads to incomplete or inconsistent scene representations.
Moreover, these methods are time-consuming during inference due to reconstruction.

To bridge this gap, our goal is to equip the MLLM with the implicit understanding of 3D structure, thereby enhancing its spatial reasoning ability.
Considering that feed-forward 3D reconstruction~\cite{wang2024dust3r, yang2025fast3r,wang2025vggt} directly predicts 3D dense structure from multi-view RGB images, these methods inherently capture the structural understanding of 3D scenes. 
To leverage this capability,  we incorporate spatial guidance into MLLM by jointly optimizing the reconstruction objective.

\noindent \textbf{Overall Architecture} \hspace{5pt} 
We build the reconstruction branch on top of Fast3R~\cite{yang2025fast3r}, which consists of a ViT encoder, a fusion transformer, and decoding heads.
Our reconstruction branch contains an encoder~$\mathcal{E}_{v}$ from MLLM, a projection layer~$\mathcal{P}$, and a reconstruction decoder~$\mathcal{D}$.
~$\mathcal{D}$ is composed of the fusion transformer and decoding heads of Fast3R.
Since Fast3R~\cite{yang2025fast3r} is optimized only for reconstruction and provides limited semantic alignment, we replace its ViT with the visual encoder from the MLLM to ensure both reconstruction and 3DVG rely on a unified representation space.
Considering that 3D reconstruction relies on dense and structure-aware features rather than semantic features, we introduce a projection layer~$\mathcal{P}$ to align and normalize the representations.
Meanwhile, we retain only the fusion transformer and decoding heads of Fast3R~\cite{yang2025fast3r}, since these blocks are responsible for multi-view geometric aggregation and pointmap prediction.

\noindent \textbf{Training Strategy} \hspace{5pt} 
Our reconstruction branch predicts the local pointmap $X_L$ and
global pointmap $X_G$. 
\begin{equation}
X_L, X_G = \mathcal{D}\bigl( \mathcal{P}(\mathcal{E}_v(I)) \bigr).
\end{equation}
For reconstruction, we adopt the confidence-weighted pointmap regression loss introduced in Fast3R~\cite{yang2025fast3r}.
During training, we optimize the reconstruction loss jointly with the other objectives required by our model.
This training strategy encourages our model to internalize 3D structure within its latent space, which ultimately enhances spatial reasoning and improves performance in 3DVG.

In practice, the reconstruction loss converges early and provides stable structure supervision, encouraging our model to acquire structure-aware features in the early stage.
It enables our model to benefit from structure understanding without compromising its multimodal alignment capability.

\subsection{Intra-view and Inter-view Attention}
\label{sec:Attention}
MLLMs cannot maintain semantic consistency across views~\cite{yeh2025seeing} and establish structure correspondences within views. For instance, the model is unable to determine whether chairs observed from different viewpoints correspond to the same physical object in 3D space.
Multi-view features naturally form two independent types of interactions: within-view and across views.
It is similar to temporal and spatial factorization in video modeling.
Inspired by~\cite{bertasius2021space}, we adopt a divided-attention design to separately capture spatial relations within each view and semantic correspondences across views.

Given multi-view image features $f\!\in\!\mathbb{R}^{B\times (V\cdot H\cdot W)\times dim}$, where $V$ is the number of views, $N{=}H\!\cdot\!W$ is the number of patches per view, and $dim$ is the feature dimension.
For each patch index $s\!\in\!\{1,\dots,N\}$, we gather patches across views $f^{\text{inter}}_{s}\!\in\!\mathbb{R}^{B\times V\times dim}$ to compute the queries, keys, and values after LayerNorm.
Inter-view attention operates across views for each patch index.
The intra-attention within each view is formulated in the same way, where patches are grouped by view index $v$ rather than by patch index $s$.

\subsection{Multi-level Position Encoding}
\label{sec:position}
Although spatial guidance encourages our MLLM to infer the 3D structure, it still lacks explicit associate visual representations with 3D positions, thereby limiting the understanding of fine-grained spatial relations. 
For example, while MLLMs can recognize objects in an image, they cannot accurately reason about their relative spatial relationships.
To address this issue, we enhance visual representations with multi-level position embeddings to incorporate explicit spatial information.

Each pixel in an RGB-D frame can be projected into a 3D coordinate. 
Inspired by NeRF~\cite{mildenhall2021nerf}, each pixel also lies on a specific camera ray.
To obtain geometry-enhanced visual representations, we encode both 3D coordinates and the camera ray’s viewing direction together with the visual embeddings.
This representation allows the model to explicitly associate visual representations with positions and viewing directions in 3D space.

Given an RGB frame, depth image $D$, camera intrinsic matrix $K \in \mathbb{R}^{3\times 3}$ and extrinsic matrix $T \in \mathbb{R}^{4\times 4}$, we calculate the 3D coordinates $p_{\text{world}} = (x,y,z)$ and camera ray’s viewing direction $\boldsymbol{\omega}$ of each pixel $(u,v)$ of the image in the global coordinate system.

Formally, for a pixel coordinate with depth value $d = D(u,v)$, the corresponding 3D point in the world coordinate system is computed as:
\begin{equation}
p_{\text{world}}
= T 
\begin{bmatrix}
d\, K^{-1}(u,v,1)^\top \\[2pt]
1
\end{bmatrix}.
\label{eq:world_coords}
\end{equation}

Each pixel corresponds to a camera ray that originates from the camera center and passes through the 3D point projected from the camera coordinate system. 
We denote a ray by $\boldsymbol{\omega} = (o,p,r)$, where $o$ is the ray origin, $p$ is the 3D point associated with the pixel, and $r$ is the viewing direction.
Formally, the ray origin in world coordinates is given by the translation matrix $t$ of the extrinsic matrix: $o_{\text{world}} = t$.
The ray termination point is equal to the 3D global coordinate $p_{\text{world}}$ of the pixel.
The normalized camera ray’s viewing direction $r$ is given by:
\begin{equation}
r = 
\frac{p_{\text{world}} - o_{\text{world}}}
{\left\lVert p_{\text{world}} - o_{\text{world}} \right\rVert_2}.
\end{equation}

After obtaining visual features $f$ from the visual encoder $\mathcal{E}_{v}$ with a small patch size, we adopt sinusoidal position encoding~$\phi( \cdot )$ to encode the average 3D coordinate $p_{\text{world}}^i = (x_i, y_i, z_i)$ of each patch $i$ in the global coordinate system following ~\cite{zheng2025video}. 
The 3D coordinate embeddings are added to the visual feature~$f_i$ of each patch as $f_i^{p}$.

Then, we perform feature aggregation via average pooling across neighboring patches to obtain a context-enriched feature $f_i^{s}$.
Lastly, we introduce a learnable positional encoding $\psi(\cdot)$ to encode the camera ray direction $r_i$ of each patch center. $\psi(\cdot)$ is implemented as a multi-layer perceptron (MLP).
The final position-aware visual representation of each patch is
\begin{equation}
f_i^{\text{vis}} 
= \text{AvgPool} \big(f_i + \phi(p_{\text{world}}^i)\big)
+ \psi(r_i).
\label{eq:multi_level_encoding}
\end{equation}

\FloatBarrier
\begin{table*}[t]
\centering
\caption{Accuracy comparison on Scanrefer~\cite{chen2020scanrefer} validation set at IoU thresholds of $0.25$ and $0.5$. We report results on the Unique subset (single-object scenes), the Multiple subset (scenes with same-class distractors), and the overall accuracy. * denotes results obtained by LoRA~\cite{hu2022lora} fine-tuning with the same parameter size as ours, while other settings follow the original paper.}
\label{tab:Scanrefer}
\resizebox{\textwidth}{!}{
\begin{tabular}{c|c|c|cc|cc|cc}
        \toprule
        \multirow{2}{*}{\textbf{Method}} & \multirow{2}{*}{\textbf{Venue}}  & \multirow{2}{*}{\textbf{LLM}} & \multicolumn{2}{c|}{\textbf{Unique}} & \multicolumn{2}{c|}{\textbf{Multiple}} & \multicolumn{2}{c}{\textbf{Overall}} 
        \\
        & & & \textbf{Acc@$\mathbf{0.25}$} & \textbf{Acc@$\mathbf{0.5}$} & \textbf{Acc@$\mathbf{0.25}$} & \textbf{Acc@$\mathbf{0.5}$} & \textbf{Acc@$\mathbf{0.25}$} & \textbf{Acc@$\mathbf{0.5}$} 
        \\
        \midrule
        ScanRefer~\cite{chen2020scanrefer} & ECCV'20 & - & $67.6$ & $46.2$ & $32.1$ & $21.3$ & $39.0$ & $26.1$
        \\
        InstanceRefer~\cite{yuan2021instancerefer}  & ICCV'21  & - & $77.5$ & $66.8$ & $31.3$ & $24.8$ & $40.2$ & $32.9$ 
        \\
        3DVG-T~\cite{zhao20213dvg}  & ICCV'21 & - & $77.2$ & $58.5$ & $38.4$ & $28.7$ & $45.9$ & $34.5$ 
        \\
        BUTD-DETR~\cite{jain2022bottom} & ECCV'22 & - & $84.2$ & $66.3$ & $46.6$ & $35.1$ & $52.2$ & $39.8$
        \\
        EDA~\cite{wu2023eda} & CVPR'23 & - & $85.8$ & $68.6$ &  $49.1$  & $37.6$  & $54.6$ &  $42.3$
        \\
        3D-VisTA~\cite{zhu20233d} & ICCV'23 & - & $81.6$ & $75.1$ & $43.7$ & $39.1$  & $50.6$ &   $45.8$ 
        \\
        VPP-Net~\cite{shi2024aware} & CVPR'24 &  - & $86.1$ & $67.1$ & $50.3$ & $39.0$  & $55.7$ &   $43.3$ 
        \\
        G3-LQ~\cite{wang2024g} & CVPR'24 &  - & $\mathbf{88.6}$ & $73.3$ & $50.2$ & $39.7$  & $56.0$ &   $44.7$ 
        \\
        MCLN~\cite{qian2024multi} & ECCV'24 & - & $86.9$ & $72.7$ & $52.0$ & $40.8$  & $57.2$ &   $45.7$ 
        \\
        ConcreteNet~\cite{unal2024four}  & ECCV'24 &  - & $86.4$ & $\mathbf{82.1}$ & $42.4$ & $38.4$  & $50.6$ &   $46.5$ 
        \\
        ViewSRD~\cite{huang2025viewsrd}  & ICCV'25 &  - & $82.1$ & $68.2$ & $37.4$ & $39.0$  & $45.4$ &   $36.0$ 
        \\
         BUTD-DETR~\cite{jain2022bottom}+AugRefer~\cite{wang2025augrefer}  & AAAI'25 &  - & $85.2$ & $69.0$ & $47.7$ & $37.2$  & $53.9$ &   $42.4$ 
        \\
         EDA~\cite{wu2023eda}+AugRefer~\cite{wang2025augrefer}  & AAAI'25 &  - & $86.2$ & $70.8$ & $50.0$ & $39.1$  & $55.7$ &   $44.0$ 
        \\
        TSP3D~\cite{guo2025tsp3d}  & CVPR'25 &  - & $87.3$ & $71.4$ & $51.0$ & $42.4$  & $56.5$ &   $46.7$ 
        \\
        \midrule
        WS-3DVG~\cite{wang2023distilling}  & ICCV'23 & - & - & - & - & -  & $27.4$ &   $22.0$ 
        \\
        \midrule
        Video-3D-LLM~\cite{zheng2025video}*  & CVPR'25 & LLaVA-Video-7B~\cite{zhang2024video}  & $82.3$ & $72.5$ & $47.2$ & $42.0$ & $54.1$  & $47.9$  
        \\
        \midrule
        LERF~\cite{kerr2023lerf} & ICCV'23 & CLIP~\cite{radford2021learning} & - & -  & - & - & $4.8$ & $0.9$ 
        \\
        OpenScene~\cite{peng2023openscene}  & CVPR'23 &  CLIP~\cite{radford2021learning}  & $20.1$ & $13.1$ & $11.1$ & $4.4$ & $13.2$ & $6.5$ 
        \\
        ZSVG3D~\cite{yuan2024visual}  & CVPR'24 & GPT-4 turbo~\cite{ouyang2022training}  & $63.8$ & $58.4$ & $27.7$ & $24.6$ & $36.4$ & $32.7$
        \\
       SeeGround~\cite{li2025seeground} & CVPR'25 &  Qwen2-VL-72B~\cite{wang2024qwen2}  & $75.7$ & $68.9$ & $34.0$ & $30.0$ & $44.1$ & $39.4$
        \\
        \midrule
        \textbf{S$^2$-MLLM} & Ours &  LLaVA-Video-7B\cite{zhang2024video} & $87.4$ & $77.8$ & $\mathbf{52.4}$ & $\mathbf{46.6}$ & $\mathbf{59.2}$ & $\mathbf{52.7}$
        \\
        \bottomrule
    \end{tabular}}
    \vspace{-2mm}
\end{table*}

\subsection{Overall Loss Function}
\label{sec:Loss}
\noindent \textbf{Visual Grounding Loss} \hspace{5pt} We follow prior work~\cite{hong20233d, huang2022multi,zheng2025video, wang2023chat, zhu20233d, zhu2024scanreason} and formulate the 3DVG task as a classification for objects proposals. 
For each bounding box \( b\) of an object in the scene, we obtain its feature \( f_{\text{obj}} \) by averaging the patch features whose projected points lie inside \( b \) with over $50\%$ coverage.
\( f_{\text{obj}} \) is further added with the 3D position embedding of the object center.
Given the hidden state \( h \) of the \(\langle\text{ground}\rangle\) token, we compute their similarity and optimize it using an InfoNCE~\cite{oord2018representation} loss. 
Formally, the grounding loss is defined as
\begin{equation}
\mathcal{L}_{\text{ground}}
= \text{InfoNCE}\big( f_{\text{obj}}, h \big),
\end{equation}
where \(\text{InfoNCE}(\cdot)\) denotes the cross-entropy contrastive objective.

\noindent \textbf{Reconstruction Loss} \hspace{5pt} Given the predicted pointmap $\hat{X}$ with confidence scores $\hat{\Sigma}$ and the ground-truth pointmap $X$, the regression loss is defined as:
\begin{equation}
\ell_{\text{regr}}(\hat{X}, X) = 
\left\lVert \tfrac{1}{\hat{z}} \hat{X} - \tfrac{1}{z} X \right\rVert_2 , 
\quad z = \tfrac{1}{|X|} \sum_{x \in X} \lVert x \rVert_2 ,
\label{eq:regr}
\end{equation}
and the final pointmap loss is formulated as:
\begin{equation}
\mathcal{L}_{X}(\hat{\Sigma}, \hat{X}, X) = 
\tfrac{1}{|X|} \sum \hat{\Sigma}_{+} \cdot 
\ell_{\text{regr}}(\hat{X}, X) 
+ \alpha \log(\hat{\Sigma}_{+}),
\label{eq:pointmap}
\end{equation}
where $\hat{\Sigma}_{+} = 1 + \exp(\hat{\Sigma})$.
Formally, the objective of reconstruction can be obtained by adding the loss of local and global point maps:
\begin{equation}
\mathcal{L}_{\text{recon}} = \mathcal{L}_{X_G} + \mathcal{L}_{X_L}.
\label{eq:recon}
\end{equation}

\noindent \textbf{Language Loss} \hspace{5pt} Visual grounding errors also arise from misclassification of the target object category~\cite{li2025seeground}, which are common in complex scenes.
For example, the model may focus on the correct spatial position but still classify a stool as a chair or confuse a nightstand with a cabinet.
To enforce semantic consistency between the predicted object category and the type of target objects, we incorporate an additional language-guidance. 
We supervise text generation using a cross-entropy loss $\mathcal{L}_{\text{lang}}$ to encourage the model produce text responses like  
``\textit{The \texttt{[object category]} is located at $\texttt{<ground>}$ in the global coordinates}''.

The final training objective combines all components:
\begin{equation}
\mathcal{L} = \lambda_{\text{g}}\mathcal{L}_{\text{ground}}
+ \lambda_{\text{r}}\mathcal{L}_{\text{recon}}
+ \lambda_{\text{l}}\mathcal{L}_{\text{lang}},
\end{equation}
where $\lambda_{\text{g}}, \lambda_{\text{r}}, \lambda_{\text{l}}$ are balancing weights.

Concretely, we construct training queries in the form of a bounding box with the object category using a unified prompt template. Please refer to the supplementary for prompt details.

\section{Experiments}

\begin{table}[t]
\centering
\caption{Accuracy comparison on Nr3D and Sr3D~\cite{achlioptas2020referit3d} validation set with both predicted at IoU thresholds of 0.25 and ground-truth bounding boxes as input. }
\label{tab:referit3d}
\resizebox{\columnwidth}{!}{\begin{tabular}{c|c|cc | cc}
        \toprule
\multirow{2}{*}{\textbf{Method}} & 
\multirow{2}{*}{\textbf{Venue}} &
\multicolumn{2}{c|}{\textbf{Pred}} & 
\multicolumn{2}{c}{\textbf{GT}} \\
\cmidrule(lr){3-4} \cmidrule(lr){5-6}
 & & \textbf{Sr3D} & \textbf{Nr3D} & \textbf{Sr3D} & \textbf{Nr3D} 
        \\
        \midrule
        InstanceRefer~\cite{yuan2021instancerefer}  & ICCV'21 &  $31.5$ & $29.9$ &  $48.0$ & $35.6$ \\
        LanguageRefer~\cite{chen2022language}  & CoRL'22  & $39.5$ & $28.6$ &  $56.0$ & $43.9$  \\
        BUTD-DETR~\cite{jain2022bottom}  & ECCV'22 &  $52.1$ & $43.3$ &  $67.0$ & $54.6$ \\
        EDA~\cite{wu2023eda}  & CVPR'23 &  $49.9$ & $40.7$ &  $68.1$ & $52.1$ \\
        MCLN~\cite{qian2024multi}  & ECCV'24 &  $\mathbf{53.9}$ & $46.1$ &  $\mathbf{68.4}$ & $\mathbf{59.8}$ \\
        \midrule
        ZSVG3D~\cite{yuan2024visual} &  CVPR'24 & $-$ & $ - $ &  $-$ & $39.0$ \\
        SeeGround~\cite{li2025seeground}  & CVPR'25 & $-$ & $-$ &  $-$ & $46.1$ 
        \\
        \midrule
        \textbf{S$^2$-MLLM} & Ours &  $\mathbf{53.9}$ & $\mathbf{50.6}$ &  $63.2$ & $\mathbf{59.8}$ 
        \\
        \bottomrule
    \end{tabular}}
    \vspace{-5mm}
\end{table}

\subsection{Dataset and Metrics}

We evaluate our method and baselines on ScanRefer~\cite{chen2020scanrefer} and ReferIt3D~\cite{achlioptas2020referit3d}.
The ReferIt3D~\cite{achlioptas2020referit3d} benchmark contains two subsets, Nr3D and Sr3D, which provide natural and synthetic referring expressions, respectively.
The ScanRefer~\cite{chen2020scanrefer} dataset includes 51,583 descriptions of 11,046 objects across ScanNet~\cite{dai2017scannet} scenes.
we train our model on the combined ScanRefer~\cite{chen2020scanrefer}, Nr3D, and Sr3D~\cite{achlioptas2020referit3d} datasets and evaluate it separately, following the setting of~\cite{zhu20233d,zheng2025video}.
This setup enables learning from both natural and template-based expressions while evaluating the model under varying levels of referential complexity.
For ScanRefer, we report accuracy at IoU thresholds of 0.25 and 0.5. 
For ReferIt3D~\cite{achlioptas2020referit3d}, we report results under two settings:
(i) using ground-truth bounding boxes (GT) as object proposals, which is a commonly used setting in previous methods~\cite{qian2024multi, jain2022bottom}, and
(ii) using predicted bounding boxes (Pred), which simulates realistic inference conditions with noisy detections.

\subsection{Comparison Methods}
For ScanRefer~\cite{chen2020scanrefer}, we compare S$^{2}$-MLLM with traditional full-supervised methods~\cite{chen2020scanrefer,yuan2021instancerefer,wu2023eda}, weakly-supervised method~\cite{wang2023distilling}, and LLM-based methods including zero-shot methods~\cite{li2025seeground} and Video-3D-LLM~\cite{zheng2025video} (for fair comparison, we also apply LoRA~\cite{hu2022lora} training on it). %which needs full parameters fine-tuning.
For ReferIt3D~\cite{achlioptas2020referit3d}, we evaluate a series of recent open-source and reproducible methods~\cite{yuan2021instancerefer,chen2022language,jain2022bottom,wu2023eda,qian2024multi,li2025seeground,yuan2024visual}. 
The results of~\cite{yuan2021instancerefer,chen2022language,jain2022bottom,wu2023eda,qian2024multi} taking predicted bounding boxes as inputs are directly taken from~\cite{guo2025tsp3d}.

\subsection{Implementation Details}
 \begin{figure*}[t]
    \centering
    \vspace{-0.5cm}
    \includegraphics[width=1.0 \linewidth]{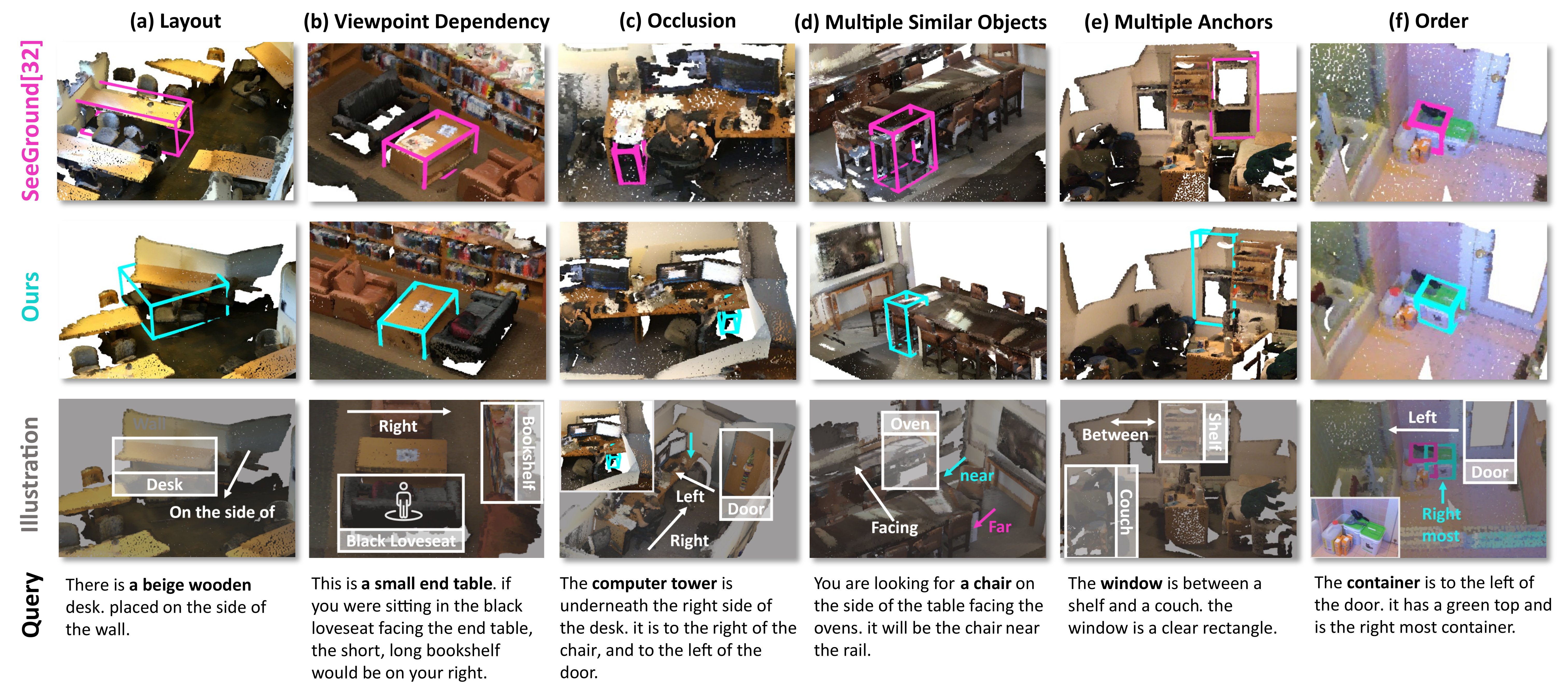}
    \vspace{-0.5cm}
    \captionof{figure}{Qualitative comparison of 3DVG results in challenging spatial understanding cases in 3DVG. Incorrect predictions are highlighted in \textcolor{magenta}{magenta}, correct ones in \textcolor{cyan}{cyan}, and key spatial relations are underlined.
Specifically, S$^2$-MLLM
(a) reasons with scene layout priors.
(b) understands the specified viewpoint. 
(c) predicts through structure cues despite partial occlusion.
(d) distinguishes similar objects by jointly reasoning over multiple spatial relationships. 
(e) accurately handles spatial relations involving multiple reference objects.
(f) understands relative positioning.
Overall, S$^2$-MLLM demonstrates more reliable spatial understanding than the previous method.} 
    \label{fig:vis}
\end{figure*}

We fine-tune our model based on LLaVA-Video~\cite{zhang2024video} 7B with LoRA~\cite{hu2022lora}. Training is performed across the combined dataset for one epoch with a batch size of 8 and a warmup ratio of $0.05$. The learning rate is scheduled to peak for the LLM and for the video encoder during warmup. All experiments are conducted only on a single A100 GPU (80GB). The temperature $\tau$ for InfoNCE loss is $0.07$.
During training, ground-truth bounding boxes are provided as object proposals.
We employ object proposals generated by Mask3D~\cite{schult2022mask3d} as predicted bounding boxes at inference. 
During training, we freeze the reconstruction decoder and fine-tune the projection layer, the visual encoder, and the language model.
During inference, the structural guidance branch is disabled.
Please refer to the supplementary for more implementation details.

\subsection{Experiment Results}
\noindent \textbf{ScanRefer}\hspace{5pt} 
Tab.~\ref{tab:Scanrefer} presents the performance of our method and prior approaches on ScanRefer~\cite{chen2020scanrefer}. 
Overall, our method achieves the best performance across all evaluation settings, reaching Acc@0.25 of $59.2\%$ and Acc@0.5 of $52.7\%$ in overall accuracy. 
Notably, our method achieves a $10.0\%$  improvement over the previous SOTA method in scenes containing multiple similar objects on Acc@0.5, which requires the model to jointly understand complex spatial relationships and accurately identify object attributes mentioned in the query (\eg, the white chair next to the desk under the window).
Meanwhile, our method improves by $4.8\%$ on Acc@0.5 compared to previous methods.
Additionally, compared with Video-3D-LLM~\cite{zheng2025video} applying LoRA-finetuning, our method improves by over $5.1\%$ on each metric. 
This improvement is attributed to our spatial guidance, which enables our model to reason about 3D layout and spatial relations, especially in complex environments.

\noindent \textbf{ReferIt3D}\hspace{5pt} 
As shown in Tab.~\ref{tab:referit3d}, our method achieves competitive results under the Pred setting, obtaining Acc@25 of $53.9\%$ on Sr3D and $50.6\%$ on Nr3D.
These gains mainly come from our spatial guidance and SE module, which produce structure-aware visual features and improve robustness to bounding-box misalignment and proposal noise.
Under the GT setting, S$^{2}$-MLLM achieves $59.8\%$ on Nr3D.
Although several methods perform better with ground-truth boxes, this idealized setting is rarely achievable in real applications.
Therefore, experimental results under the predicted-box setting reflect real-world performance more accurately.
The relatively lower performance on Sr3D under the GT setting is mainly due to the template-based queries.
Traditional fully supervised methods can easily exploit these fixed patterns to match them with object features.
In contrast, our performance on Nr3D highlights the advantage of S$^{2}$-MLLM in understanding natural human queries and aligning them with 3D spatial information.

\noindent \textbf{Qualitative Results}\hspace{5pt} 
To further demonstrate the effectiveness of our approach, we visualize the visual grounding results of S$^2$-MLLM and SeeGround~\cite{li2025seeground}.
As shown in Fig.~\ref{fig:vis}, S$^{2}$-MLLM demonstrates clear advantages in challenging spatial reasoning scenes, including layout understanding, viewpoint specification, occlusion, and grounding among multiple similar objects.
The visualization results indicate that S$^2$-MLLM can capture 3D structure and understand complex spatial relations, indicating the effectiveness of spatial guidance.

\begin{table}[t]
\centering
\small
\setlength{\tabcolsep}{4pt}
\caption{
\textbf{Efficiency comparison.}
We report the training cost (in GPU hours), trainable parameters (in MB), and the inference latency (in seconds). $t_0$ represents the additional inference time of reconstructing point clouds.
}
\resizebox{\columnwidth}{!}{\begin{tabular}{c|ccc}
\toprule
\makecell{\textbf{Method}} &
\makecell{\textbf{GPU}\\\textbf{Hours} $\downarrow$} &
\makecell{\textbf{Trainable}\\\textbf{Parameters (MB)}$\downarrow$} &
\makecell{\textbf{Latency}\\\textbf{(s)}$\downarrow$} \\
\midrule
Video-3D-LLM~\cite{zheng2025video}  & $256$ & $8078.79$ & $1.04$  \\
SeeGround~\cite{li2025seeground} & - & - &  $3.97+t_0$\\
S$^2$-MLLM(w/o SG)              & $65$ & $1763.53$ & $1.16$  \\
S$^2$-MLLM(Full)                 & $72$ & $1767.50$ & $1.16$  \\
\bottomrule
\end{tabular}}

\label{tab:efficency}
\vspace{-5mm}
\end{table}
\subsection{Efficiency and Generalization}
\noindent \textbf{Efficiency}\hspace{5pt} 
Tab.~\ref{tab:efficency} compares the training efficiency and inference latency of our method and ~\cite{zheng2025video,li2025seeground}.
All experiments are conducted on a single A100 GPU (80G). 
Due to limited computational resources, we utilize Qwen2-VL-7B~\cite{wang2024qwen2} to measure the inference latency of SeeGround~\cite{li2025seeground}.
Our S$^2$-MLLM requires only $25\%$ of the trainable parameters and GPU hours required by the full-parameter fine-tuning method~\cite{zheng2025video}, while achieving higher accuracy.
Although zero-shot methods are training-free, they typically perform multiple API calls within a single inference, leading to nearly $4\times$ higher inference time of SeeGround~\cite{li2025seeground}.
Notice that SeeGround~\cite{li2025seeground} needs extra inference time $t_0$ for point clouds reconstruction, which further increases the actual latency.
We further analyze the overhead introduced by our proposed spatial guidance. 
SG adds only around $10\%$ training time, with negligible additional trainable parameters.
During inference, S$^2$-MLLM avoids extra point-cloud reconstruction and multi-view rendering, meaning that SG introduces no additional inference latency.
Combining the performance in Tab.~\ref{tab:Scanrefer} and Tab.~\ref{tab:referit3d} with the efficiency comparison, our method achieves the best trade-off between performance and efficiency, demonstrating strong potential for real-world applications.

\noindent \textbf{Out-Of-Distribution Dimensions}\hspace{5pt} 
We evaluate out-of-distribution (OOD) performance on MultiScan~\cite{mao2022multiscan} and ArtiScenes~\cite{baruch2021arkitscenes}, comparing S$^{2}$-MLLM with a traditional supervised method MCLN~\cite{qian2024multi} and a zero-shot method SeeGround~\cite{li2025seeground}.
We train our model on the union datasets of ScanRefer~\cite{chen2020scanrefer} and ReferIt3D~\cite{achlioptas2020referit3d}, which are built on ScanNet~\cite{dai2017scannet}. 
Then we directly evaluate the OOD performance of our model without any fine-tuning.
Compared with ScanNet~\cite{dai2017scannet}, MultiScan~\cite{mao2022multiscan} and ArtiScenes~\cite{baruch2021arkitscenes} introduce substantial distribution shifts in scene layouts, object compositions, and language descriptions.
As shown in Table~\ref{tab:ood}, our S$^2$-MLLM achieves the best performance under both OOD benchmarks with Acc@25 of $59.13$ on Multiscan~\cite{mao2022multiscan} and $43.26$ on ArkiScenes~\cite{baruch2021arkitscenes}.
The results show that our model acquires spatial reasoning ability through spatial guidance rather than learning the distribution of a specific dataset.

\begin{table}[t]
\centering
\small
\setlength{\tabcolsep}{4pt}
\caption{
\textbf{Out-of-Distribution (OOD) Evaluation} on Multiscan~\cite{mao2022multiscan} and ArkiScenes~\cite{baruch2021arkitscenes}.
}
\resizebox{\columnwidth}{!}{\begin{tabular}{c|cc|cc}
\toprule
\multirow{2}{*}{\textbf{Method}} & \multicolumn{2}{c|}{\textbf{MultiScan}} & \multicolumn{2}{c}{\textbf{ArtiScenes}} 
\\
  & \textbf{Acc@$\mathbf{0.25}$} & \textbf{Acc@$\mathbf{0.5}$} & \textbf{Acc@$\mathbf{0.25}$} & \textbf{Acc@$\mathbf{0.5}$} 
  \\
\midrule
MCLN~\cite{qian2024multi}  & 12.91 & 6.00 & 17.21 & 6.35  \\
SeeGround~\cite{li2025seeground} & 53.41 & 53.41 & 38.82 & 38.43  \\
S$^2$-MLLM & \textbf{59.13} & \textbf{53.62} & \textbf{43.26} & \textbf{39.84}  \\
\bottomrule
\end{tabular}}
\label{tab:ood}
\vspace{-0.05in}
\end{table}

\begin{table}[t]
\centering
\small
\setlength{\tabcolsep}{3pt}
\caption{
\textbf{Ablation study} on the ScanRefer~\cite{chen2020scanrefer} dataset.
We evaluate the contribution of each proposed component and the impact of the number of input frames.
(SG) Spatial Guidance; (MPE) Multi-level Position Encoding;
(Attn) Intra-view and Inter-view Attention; (LG) Language Guidance.
}
\begin{tabular}{ c|c| cc }
\toprule
\textbf{Ablation} &  \textbf{Num Frames} & \textbf{Overall@0.25} & \textbf{Overall@0.5}  \\
\midrule
\multirow{2}{*}{w/o SG} 
& $16$ & 54.40 & 48.45 \\
 & $24$ & 56.46 & 49.88 \\
\midrule
w/o MPE & $16$ & 44.13 & 38.49 \\
w/o Attn & $16$ & 59.13 & 52.30 \\
w/o LG & $16$ & 57.75 & 50.85 \\
\midrule
\multirow{2}{*}{Full (S$^{2}$-MLLM)} 
 &$16$ & \textbf{59.18} & \textbf{52.67} \\
 &$24$ & \textbf{60.59} & \textbf{53.66} \\
\bottomrule
\end{tabular}

\label{tab:ablation}
\end{table}

\subsection{Ablation}

\noindent \textbf{Ablation on Our Modules}\hspace{5pt}  
We validate the effectiveness of each component in S$^{2}$-MLLM.
As shown in Tab.~\ref{tab:ablation}, all components bring clear performance gains. 
Removing spatial guidance (SG) leads to a clear drop of $4.78\%$ Acc@0.25 and $4.22\%$ Acc@0.5 under the 16 frames setting, demonstrating the importance of enforcing structure supervision.
Removing multi-level positional encoding (MPE) causes the largest degradation of $15.05\%$ at Acc@0.25, indicating that explicit position information is crucial for LLM-based 3DVG.

\noindent \textbf{Ablation on Frames.}\hspace{5pt} 
We further conduct ablation studies on the effect of the number of input frames.
Increasing frames from $16$ to $24$ consistently improves performance across all settings.
The gains are more pronounced in the setting without spatial guidance (SG), reaching $+2.06\%$ at Acc@0.25 and $+1.43\%$ at Acc@0.5.
In contrast, the improvement becomes marginal with spatial guidance enabled ($+1.41\%$ at Acc@0.25 and $+0.99\%$ at Acc@0.5), while GPU memory usage and training time increase significantly.
This comparison demonstrates that spatial guidance already enables the model to extract reliable structure cues and perform spatial reasoning even from sparse observations, reducing the dependence on dense multi-view inputs.
This highlights the effectiveness of SG in enhancing spatial understanding while keeping computational overhead low.

\section{Conclusion}
In this work, we propose S$^{2}$-MLLM, a novel framework that equips MLLMs with implicit 3D reasoning capability for 3D visual grounding. By integrating feed-forward reconstruction to provide spatial guidance and introducing a structure-enhanced module, our model is enable to understand 3D scenes and perform latent spatial reasoning without requiring explicit point-cloud reconstruction at inference. Extensive experiments on both in-domain and out-of-domain benchmarks evaluate the performance of S$^{2}$-MLLM. The results verify that S$^{2}$-MLLM achieves outstanding accuracy, efficiency, and generalization, demonstrating its potential for real-world embodied applications.

\clearpage
{
    \small
    \bibliographystyle{ieeenat_fullname}
    \bibliography{main}
}

\clearpage

\appendix
\clearpage
\setcounter{page}{1}
\maketitlesupplementary

\section{Implementation Details}
\label{sec:Details}
\subsection{Training Details}
During training, we freeze the reconstruction decoder and fine-tune the projection layer, the visual encoder, and the language model.
Since the LLM output resides in the text space, whereas 3D visual grounding requires regression to region-level features, using LoRA~\cite{hu2022lora} alone is insufficient for learning cross-modal alignment. 
Therefore, we additionally fully finetune the last four layers of the LLM, the projection layer, as well as the inter-view and intra-view attention during training.
The rest of the model is fine-tuned using LoRA~\cite{hu2022lora}, including the visual encoder.

\subsection{Dataset and Arguments}
Following~\cite{zhu20233d,zheng2025video}, we train our model on the combined dataset of ScanRefer~\cite{chen2020scanrefer}, Nr3D~\cite{achlioptas2020referit3d}, and Sr3D~\cite{achlioptas2020referit3d}, and evaluate it separately on each corresponding validation set. 
We provide detailed statistics about the data used for training and evaluation, including out-of-distribution (OOD) evaluation in Tab.~\ref{tab:supp_data}. Following~\cite{zheng2025video}, we convert all data into the LLaVA~\cite{liu2023visual} format and report statistics based on this unified format.

In addition, we show detailed hyperparameters in our experiments in Tab.~\ref{tab:supp_hyper}
\begin{table}[h]
\centering
\small
\setlength{\tabcolsep}{4pt}

\caption{Detailed statistics of datasets.}
\label{tab:supp_data}

\resizebox{\columnwidth}{!}{
\begin{tabular}{c|c| c c c}
\toprule
\textbf{Split} & \textbf{Dataset} & \textbf{Samples} & \textbf{Scenes} & \textbf{Query Length} \\
\midrule
\multirow{3}{*}{Train} 
& ScanRefer~\cite{chen2020scanrefer} & 36665 & 562 & 17.83 \\
& Nr3D~\cite{achlioptas2020referit3d} & 32919 & 511 & 25.38 \\
& Sr3D~\cite{achlioptas2020referit3d} & 65844 & 1018 & 23.68 \\
\midrule

\multirow{3}{*}{Test} 
& ScanRefer~\cite{chen2020scanrefer} & 9508 & 141 & 17.92 \\
& Nr3D~\cite{achlioptas2020referit3d} & 8584 & 130 & 25.12 \\
& Sr3D~\cite{achlioptas2020referit3d} & 17726 & 255 & 23.72 \\
\midrule

\multirow{2}{*}{OOD}
& MultiScan~\cite{mao2022multiscan} & 1490 & 53 & 20.44 \\
& ArkitScenes~\cite{baruch2021arkitscenes} & 2693 & 275 & 20.88 \\
\bottomrule
\end{tabular}
}
\end{table}

\begin{table}[htbp]
    \centering
    \caption{Detailed statistics of hyperparameters.}
    \begin{tabular}{c|c}
    \toprule
       Parameter Name &  Value \\
    \midrule
      \multicolumn{2}{c}{Training} \\
      \midrule
       LoRA rank  &  64 \\
       LoRA $\alpha$ &  16 \\
       LoRA dropout & 0.05 \\
       LoRA bias & None \\
       FP/BF Precision & bf16 \\
       tf32 & False \\
       weight Decay & 0.0 \\
       tuning MLP or ViT & True \\
       training steps & 16928 \\
       batch size & 1 \\
       warmup ratio & 0.05 \\
       lr & 2e-5 \\
       lr of ViT & 2e-4 \\
       optimizer & AdamW \\
       max token length & 32768 \\
       gradient accumulation steps & 8 \\
       $\lambda_g$ & 1.0 \\
       $\lambda_r$ & 0.3 \\
       $\lambda_l$ & 1.0 \\
       \midrule
       \multicolumn{2}{c}{Inference} \\
       \midrule 
      temperature & 1.0 \\
      num beams & 1 \\
      top\_p, top\_k & 1.0, 50 \\
    \bottomrule
    \end{tabular}
    \label{tab:supp_hyper}
\end{table}

\section{Additional Ablation Studies}
\begin{table}[t]
\centering
\small
\setlength{\tabcolsep}{3pt}
\caption{
\textbf{Ablation study} on the ScanRefer~\cite{chen2020scanrefer} dataset.
We evaluate the contribution of inter-view and intra-view attention (Attn).
}
\resizebox{0.3\textwidth}{!}{\begin{tabular}{ c|cc }
\toprule
\textbf{Ablation} &  \textbf{Overall@$\mathbf{0.25}$} &  \textbf{Overall@$\mathbf{0.5}$} \\

\midrule
\textbf{Base} & $5.31$ & $5.07$ \\
\textbf{Base+Attn} & $\mathbf{41.74}$ & $\mathbf{35.78}$   \\
\bottomrule
\end{tabular}}

\label{tab:sup_ab}
\end{table}

To further verify the contribution of the inter-view and intra-view attention (Attn), we provide an additional ablation experiment.  
Specifically, we report results for LLava-Video 7B~\cite{zhang2024video} (Base) and Base + Attn, which adds only Attn.  
This comparison enables a clearer evaluation of the effect of Attn, isolated from the influence of other components.
As shown in Tab.~\ref{tab:sup_ab}, the inclusion of Attn significantly improves the performance of S$^2$-MLLM in 3DVG. 
Since other modules we proposed are highly effective and contribute substantially to the overall performance, removing the Attn module in the ablation study mentioned in Tab.~\ref{tab:ablation} of the main text does not result in a significant performance drop.

\section{Additional Analysis}
 \begin{figure}[t]
    \centering
    \includegraphics[width=1.0 \linewidth]{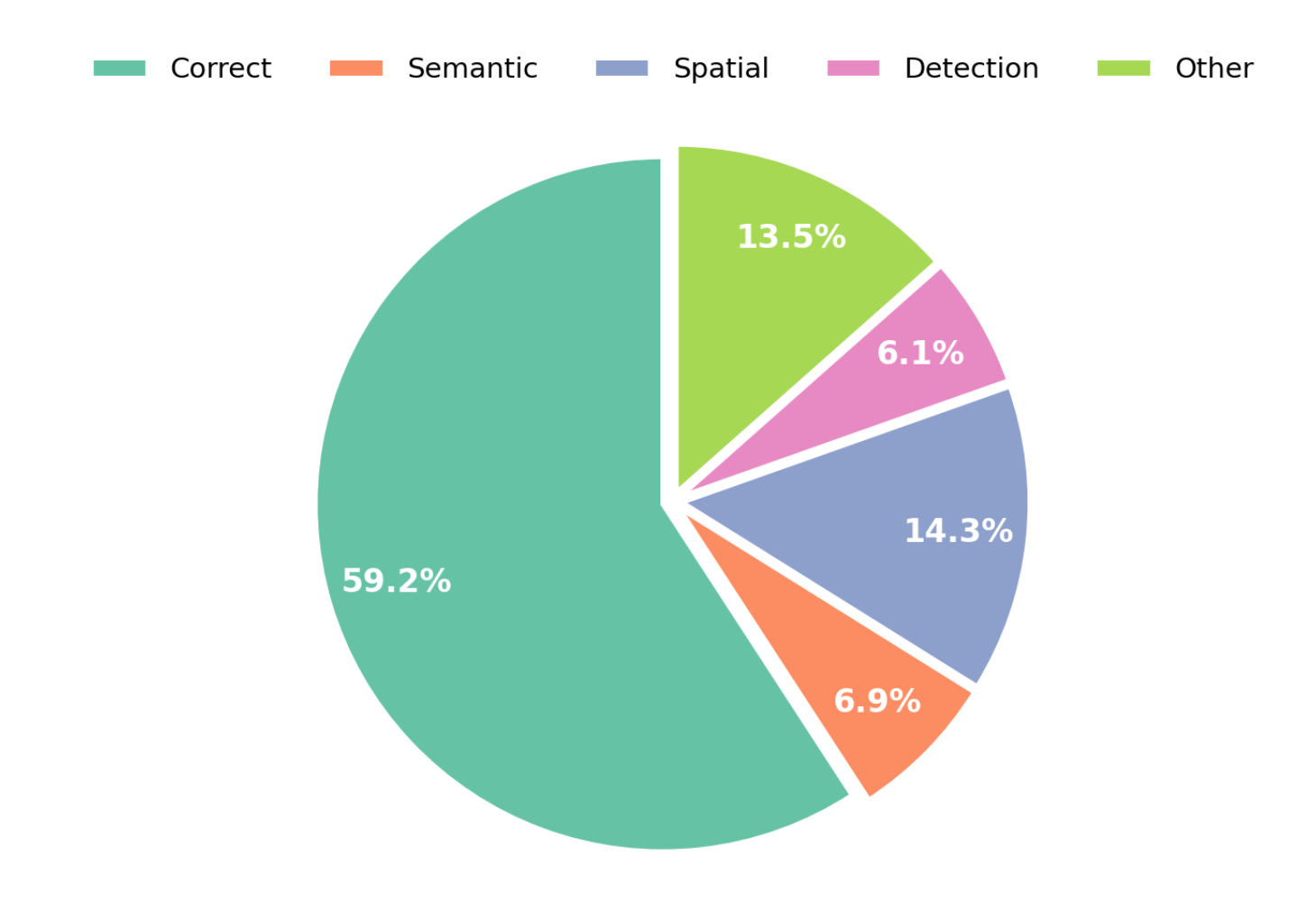}
    \captionof{figure}{Error type analysis on ScanRefer~\cite{chen2020scanrefer} dataset.} 
    \label{fig:error}
\end{figure}

 \begin{figure*}[t]
    \centering
    \includegraphics[width=1.0 \linewidth]{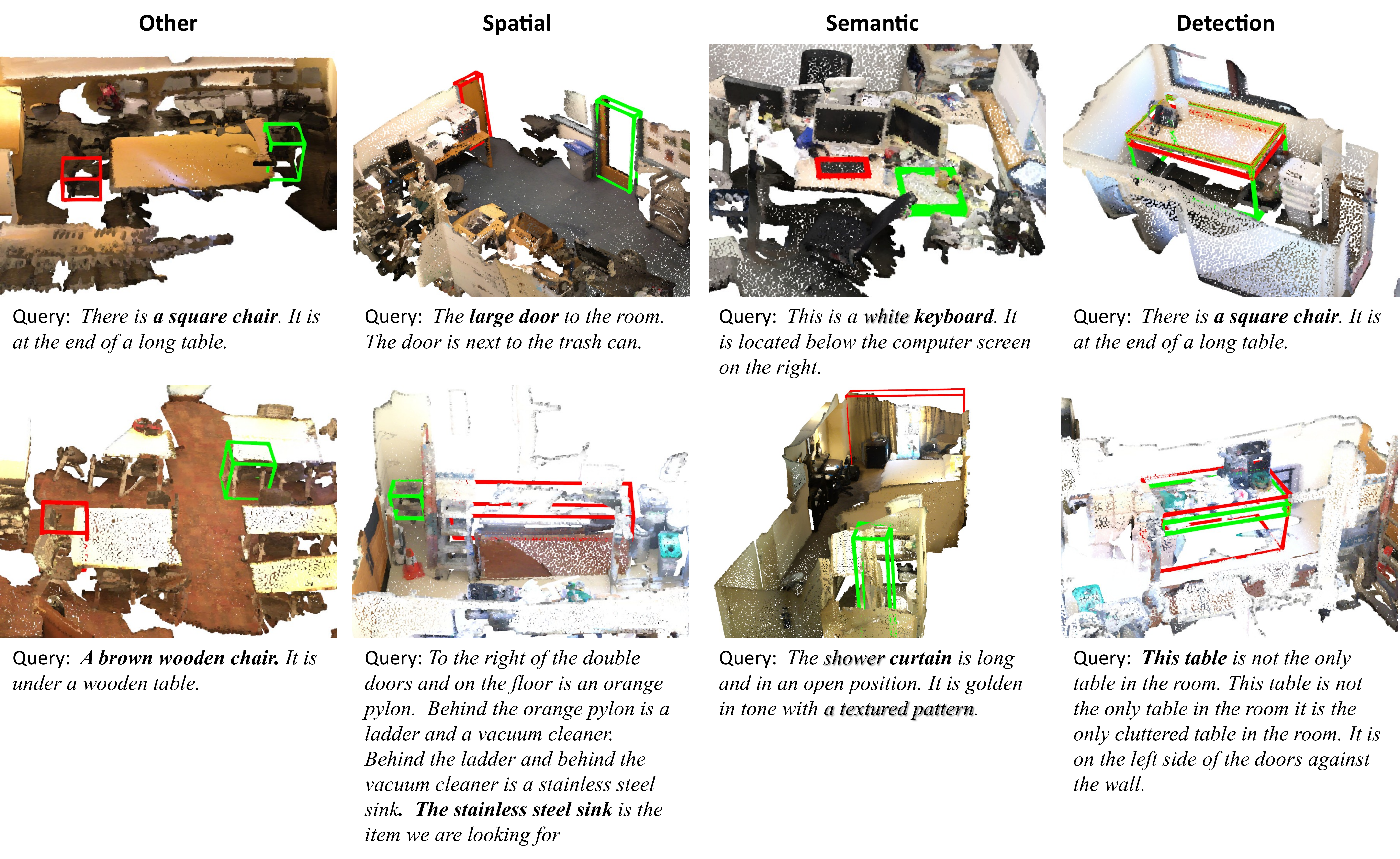}
    \captionof{figure}{Typical types of failure cases in Scanrefer~\cite{chen2020scanrefer}. Ground Truth is highlighted in \textcolor{green}{green}, our predictions in \textcolor{red}{red}.  Key semantic information is in shadow.} 
    \label{fig:err_case}
\end{figure*}

We analyzed the error types in the predictions of S$^2$-MLLM in Fig.~\ref{fig:error} and visualize typical error cases in Fig.~\ref{fig:err_case}.
We adopted the same definition of error types as~\cite{jin2025spazer}, classifying errors into four types.
Overall, S$^2$-MLLM achieves an accuracy of $59.2\%$ on ScanRefer~\cite{chen2020scanrefer}, indicating that our model can understand language descriptions and 3D scenes in most cases.

Among the errors, we observed that:
(1) Spatial: As shown in Fig.~\ref{fig:err_case}, these cases mainly involve complex relational descriptions with multiple anchor objects (\eg, determining the target object based on multiple anchor objects or complex relational descriptions that require multi-step reasoning).
(2) Semantic: These errors mainly occur due to misjudgment of fine-grained attributes of the target object and anchor objects (\eg, the color of the keyboard, the pattern and usage of the curtain).
(3) Detection: These errors arise because the bounding boxes provided by the detector are inaccurate. 
In these cases, detectors usually predict the correct object but predict an inaccurate 3D bounding region, often due to occlusion, partial visibility, or sparse viewpoints.        
This suggests that improving the robustness under limited view coverage could further enhance performance.
(4) Other: This category primarily refers to inaccurate language descriptions, where the predicted object matches the description but differs from the ground truth. This indicates that current 3DVG datasets still have limitations and incompleteness, rather than being due to the model’s performance.

\section{Additional Qualitative Results}
\subsection{Visualizations on Nr3D}
 \begin{figure*}[t]
    \centering
    \includegraphics[width=1.0 \linewidth]{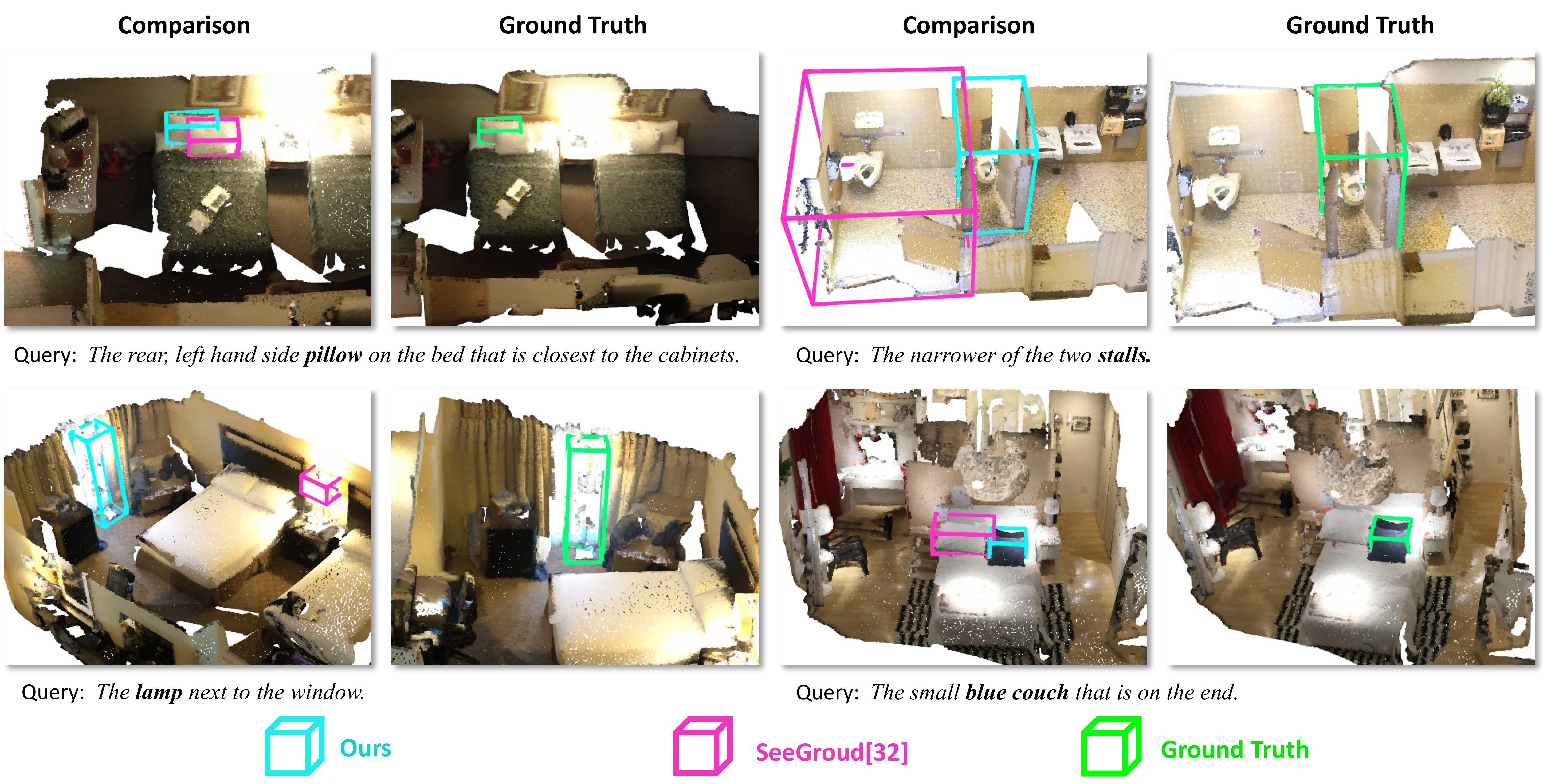}
    \captionof{figure}{Qualitative comparison of 3DVG results in Nr3D~\cite{achlioptas2020referit3d}. Ground Truth is highlighted in \textcolor{green}{green}, our predictions in \textcolor{cyan}{cyan}, and predictions of SeeGround~\cite{li2025seeground} in \textcolor{magenta}{magenta}.} 
    \label{fig:nr3d}
\end{figure*}

We additionally visualize the performance of S$^2$-MLLM and SeeGround~\cite{li2025seeground} on Nr3D~\cite{achlioptas2020referit3d}, further demonstrating the advantages of our approach. 
As shown in Fig.~\ref{fig:nr3d}, S$^2$-MLLM is capable of accurately understanding the language descriptions and the spatial relationships in the 3D scene, such as the relative size of two similar regions, the distance between two similar objects, and the window.
This demonstrates that our model exhibits superior 3D spatial understanding and reasoning abilities.

\subsection{Ablation Visualizations}
 \begin{figure*}[t]
    \centering
    \includegraphics[width=0.8 \linewidth]{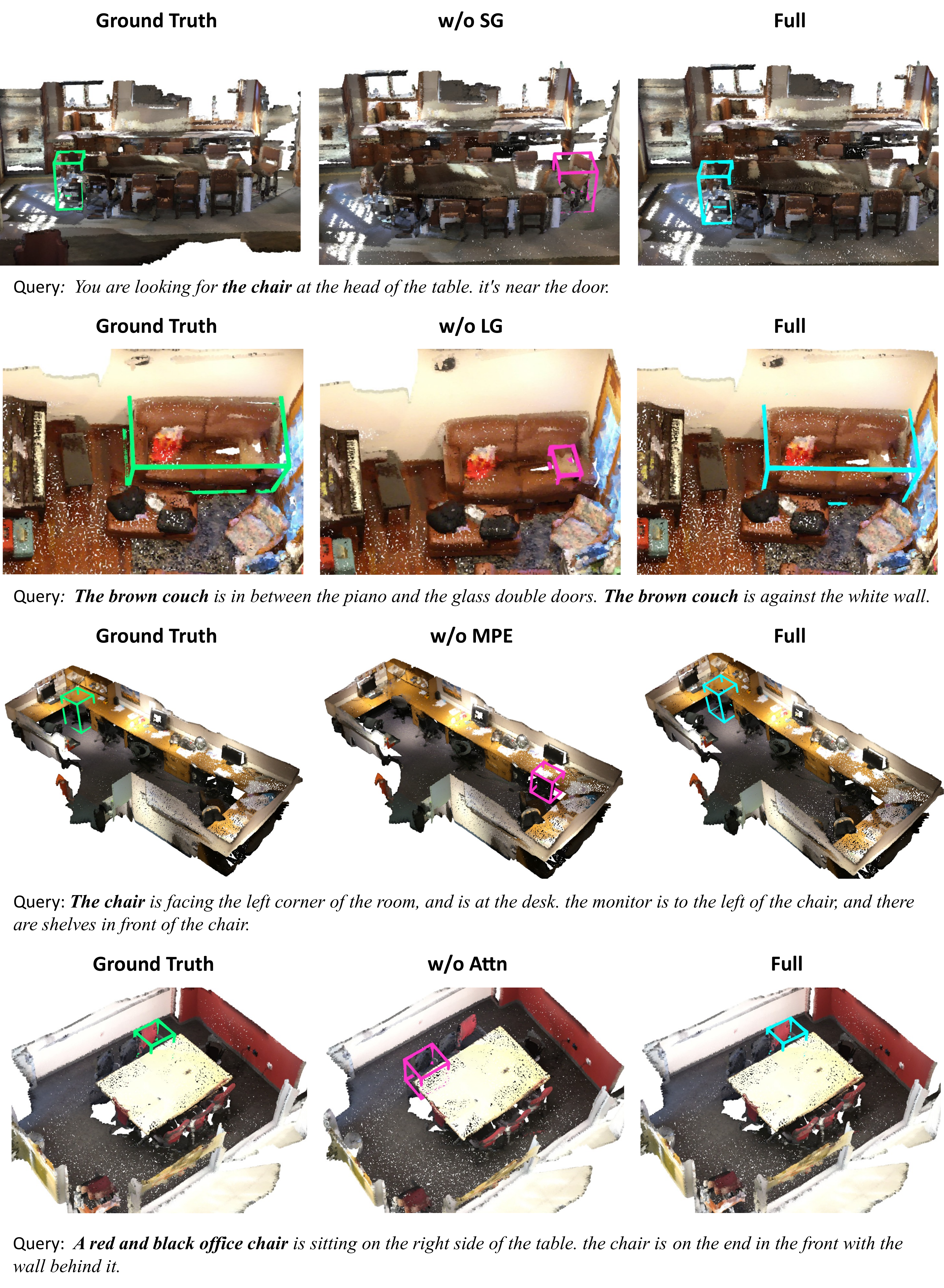}
    \captionof{figure}{Qualitative ablation results in Scanrefer~\cite{chen2020scanrefer}. Ground Truth is highlighted in \textcolor{mygreen}{mygreen}, predictions of our full model are in \textcolor{cyan}{cyan}, and predictions of the model without specific module are in \textcolor{magenta}{magenta}.  (SG) Spatial Guidance; (MPE) Multi-level Position Encoding;
(Attn) Intra-view and Inter-view Attention; (LG) Language Guidance.} 
    \label{fig:ablation}
\end{figure*}

We provide additional visual examples on the ScanRefer~\cite{chen2020scanrefer} to supplement the ablation analysis, further illustrating the effectiveness of each component we proposed in S$^2$-MLLM.
As shown in Fig.~\ref{fig:ablation}, each row corresponds to different ablation experiments.
We observe that spatial guidance (SG) enables S$^2$-MLLM to accurately comprehend the spatial relationship (\eg, the chair at the head of the table).
The second row highlights the importance of LG: without language guidance (LG), S$^2$-MLLM fails to identify the brown couch correctly in the context of surrounding objects.
With multi-level position encoding (MPE), encoding camera rays' viewing directions encourages S$^2$-MLLM to distinguish orientation and observation direction (\eg, identifying the chair facing the desk).
The fourth row shows the impact of inter-view and intra-view attention (Attn): S$^2$-MLLM without Attn struggles to locate the red and black office chair at the correct position relative to the table, while the full model successfully identifies it.
In the presence of Attn, S$^2$-MLLM can stably identify the target object even when the query involves viewpoint transitions (\eg, describing the target object from different perspectives or based on different anchor objects).
These results demonstrate how each part we proposed contributes to improving the ability of S$^2$-MLLM to understand and reason in the 3D scenes.
\subsection{Out-of-Distribution Qualitative Examples}
 \begin{figure*}[t]
    \centering
    \includegraphics[width=1.0 \linewidth]{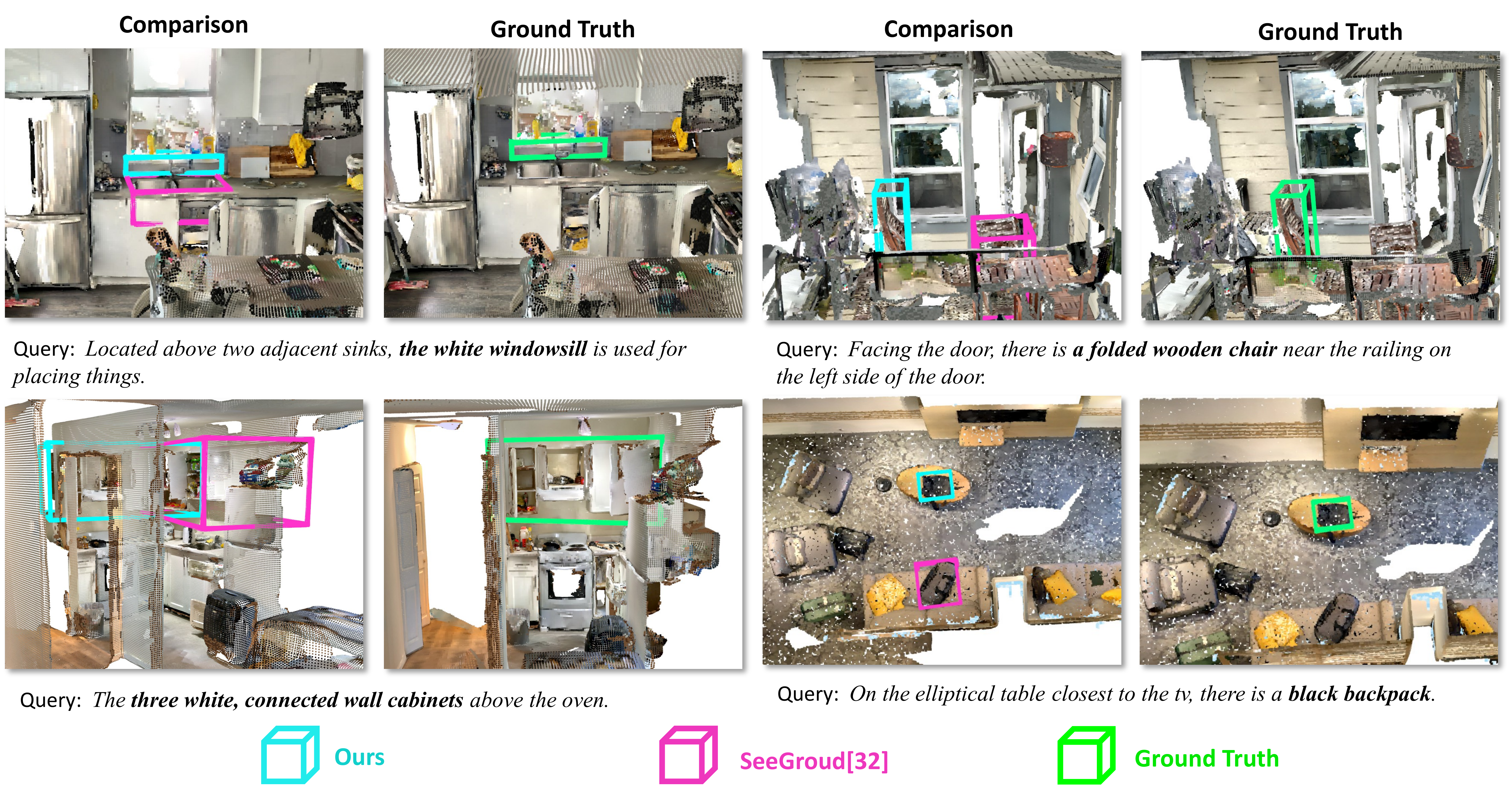}
    \captionof{figure}{Qualitative comparison of 3DVG results in Multiscan~\cite{mao2022multiscan}. Ground Truth is highlighted in \textcolor{green}{green}, our predictions in \textcolor{cyan}{cyan}, and predictions of SeeGround~\cite{li2025seeground} in \textcolor{magenta}{magenta}. } 
    \label{fig:multiscan}
\end{figure*}

 \begin{figure*}[t]
    \centering
    \includegraphics[width=1.0 \linewidth]{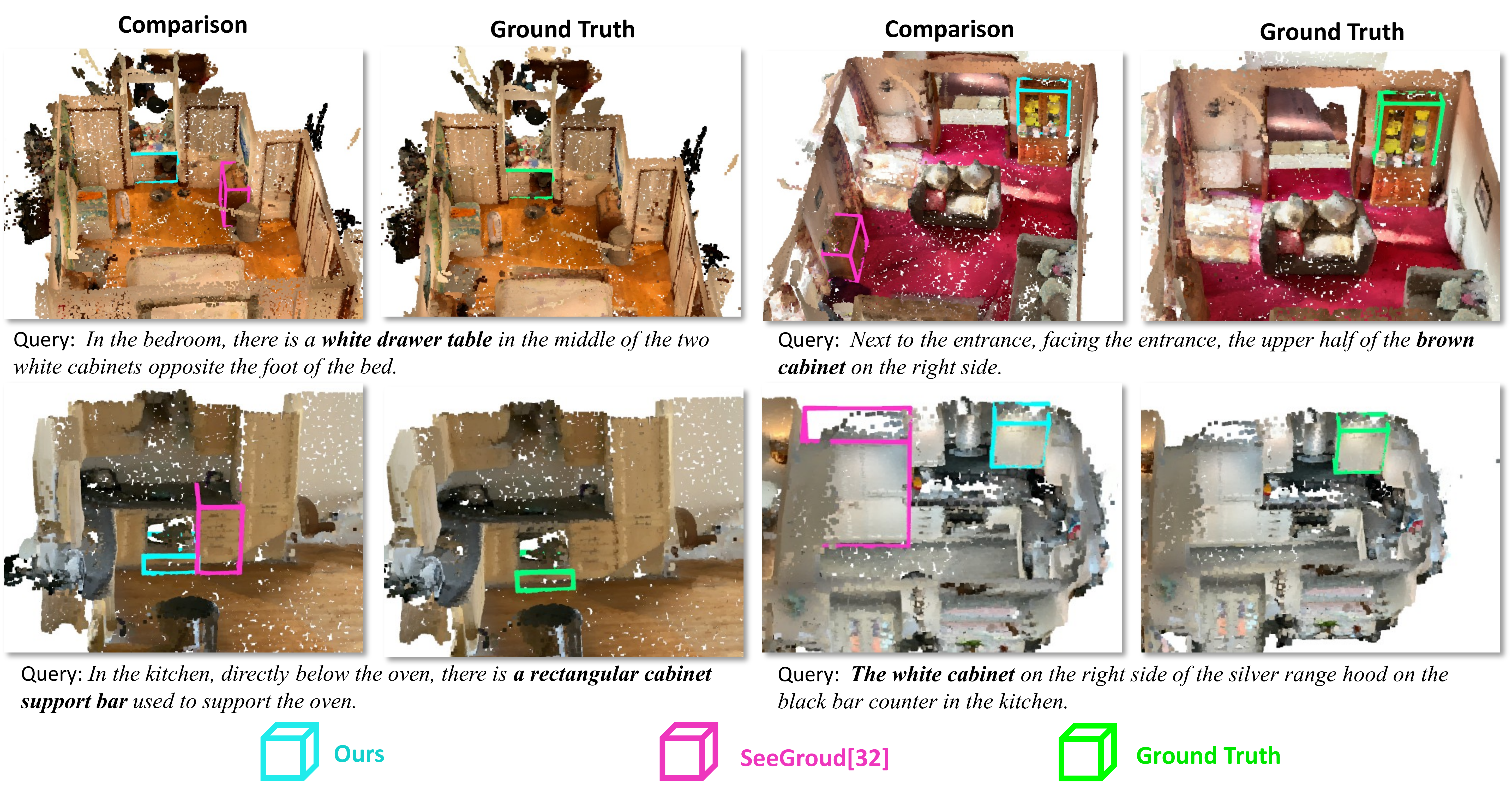}
    \captionof{figure}{Qualitative comparison of 3DVG results in ArkiScenes~\cite{baruch2021arkitscenes}. Ground Truth is highlighted in \textcolor{green}{green}, our predictions in \textcolor{cyan}{cyan}, and predictions of SeeGround~\cite{li2025seeground} in \textcolor{magenta}{magenta}.} 
    \label{fig:arki}
\end{figure*}

Fig.~\ref{fig:multiscan} and Fig.~\ref{fig:arki} provide additional qualitative examples on Multiscan~\cite{mao2022multiscan} and Arkiscenes~\cite{baruch2021arkitscenes} compared with SeeGroud~\cite{li2025seeground}.
As shown in Fig.~\ref{fig:multiscan} and Fig.~\ref{fig:arki}, the scene layouts of MultiScan~\cite{mao2022multiscan} and ArkiScenes~\cite{baruch2021arkitscenes} differ from those in ScanNet~\cite{dai2017scannet}. 
Despite being completely out of distribution, S$^2$-MLLM is still able to maintain strong spatial reasoning capabilities and achieves superior performance.

% To split the supplementary pages from the main paper, you can use \href{https://support.apple.com/en-ca/guide/preview/prvw11793/mac#:~:text=Delete%20a%20page%20from%20a,or%20choose%20Edit%20%3E%20Delete).}{Preview (on macOS)}, \href{https://www.adobe.com/acrobat/how-to/delete-pages-from-pdf.html#:~:text=Choose%20%E2%80%9CTools%E2%80%9D%20%3E%20%E2%80%9COrganize,or%20pages%20from%20the%20file.}{Adobe Acrobat} (on all OSs), as well as \href{https://superuser.com/questions/517986/is-it-possible-to-delete-some-pages-of-a-pdf-document}{command line tools}.
\end{document}